%% file: main.tex
\documentclass[10pt,twocolumn,letterpaper]{article}

\usepackage{cvpr}              
\usepackage{CJKutf8}

\usepackage{float}

\input{preamble}

\definecolor{cvprblue}{rgb}{0.21,0.49,0.74}
\usepackage[pagebackref,breaklinks,colorlinks,citecolor=cvprblue]{hyperref}

\def\paperID{*****} 
\def\confName{CVPR}
\def\confYear{2024}

\title{Perceptual Similarity guidance and text guidance optimization for Editing Real Images using Guided Diffusion Models}

\author{Ruichen Zhang\\
\small School of Mechanical, Electrical and Information Engineering, Shandong University\\
{\tt\small ruichenzhangedu@gmail.com}
}

\begin{document}
\begin{CJK}{UTF8}{gbsn}
\twocolumn[{%
\renewcommand\twocolumn[1][]{#1}%
\maketitle
\begin{center}
    \centering
    \captionsetup{type=figure}
    \includegraphics[width=1\textwidth,height=7cm]{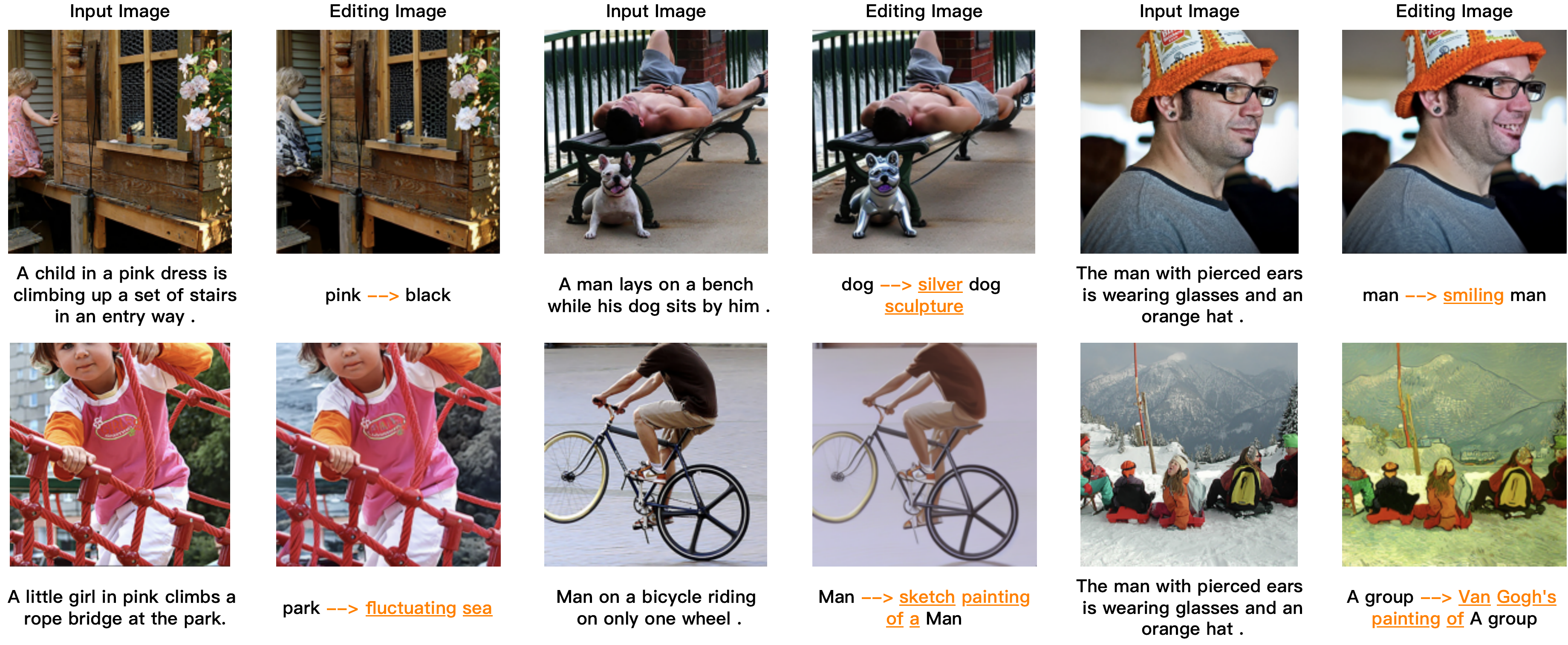}
    
    \captionof{figure}{\textbf{Perceptual Similarity guidance and text guidance optimization for real image editing.} Our method edits the picture by taking the real images with the image caption prompt and the edited prompt as input.}
\end{center}%
}]

\input{sec/0_abstract}    
\input{sec/1_intro}

\input{sec/2_formatting}

\input{sec/3_finalcopy}

\input{sec/4_Experiment}

\input{sec/5_futurework}

{
    \small
    \bibliographystyle{ieeenat_fullname}
    \bibliography{main}
}

\end{CJK}
\end{document}

%% file: preamble.tex
\usepackage[dvipsnames]{xcolor}


%% file: sec/0_abstract.tex
\begin{abstract}
When using a diffusion model for image editing, there are times when the modified image can differ greatly from the source. To address this, we apply a dual-guidance approach to maintain high fidelity to the original in areas that are not altered. First, we employ text-guided optimization, using text embeddings to direct latent space and classifier-free guidance. Second, we use perceptual similarity guidance, optimizing latent vectors with posterior sampling via Tweedie’s formula during the reverse process. This method ensures the realistic rendering of both the edited elements and the preservation of the unedited parts of the original image.
\end{abstract}

%% file: sec/1_intro.tex
\section{Introduction}
\label{sec:intro}
The text-guided stable diffusion model \cite{hertz2022prompttoprompt} has garnered significant attention due to its authenticity and precision. Recent studies, employing classifier-free guidance and DDIM inversion \cite{mokady2022nulltext} , offer viable approaches for real image editing. Nevertheless, there is room for improvement in the image guidance generation process to enhance the method's ability to bring the generated image closer to the given prompt and the original image.

Whether it involves editing the image by blending forward and backward images with an adjustable mixing ratio \cite{choi2021ilvr} , employing masks for local modifications \cite{lugmayr2022repaint}, or preserving specific low-frequency information \cite{meng2022sdedit}, a common issue arises with the relatively limited applicability of these approaches. In classifier-free guidance \cite{ho2022classifierfree}, excessive emphasis is often placed on distinguishing between the original prompt and the unconditional prompt, neglecting effective consideration of the distance between the original prompt and the target prompt. Additionally, many methods primarily rely on prompts to initialize latent features, rather than incorporating the distance between the generated image and the real image at the image level.

In this paper, we present an effective bootstrap method aimed at enhancing the consistency of image editing with the original image. Our approach revolves around guided calculations involving two pivotal facets of the guided diffusion model: Perceptual Similarity \cite{zhang2018unreasonable} guidance and optimization of text guidance.

In the optimization of text guidance, three predictions are generated at each diffusion step: one using unconditional text embeddings, another with the image caption prompt as the condition, and a third with an editing text prompt as the condition. We applied weights to these three predictions to amplify the impact of edited text on the image while preserving similarity with the real image.

In Perceptual Similarity guidance, we conduct a posterior sample to estimate $P(Z_0 | Z_t)$ utilizing Tweedie's formula. At each reverse step, we decode the latent variables back into the image domain and employ Perceptual Similarity to compute the pixel-level difference between the edited image and the real image for guidance.


%% file: sec/2_formatting.tex
\section{Related Work}
\label{sec:formatting}

The field of image editing has witnessed remarkable advancements in recent years, providing a plethora of potent tools and techniques to enhance the efficiency and creativity of image processing. This encompasses various functions such as image restoration and enhancement \cite{yang2022adaint}, image style conversion \cite{park2023lanit}, face editing \cite{gao2021highfidelity}, super resolution \cite{liang2022details}, among others. The advent of the text-guided stable diffusion model \cite{sheynin2022knndiffusion,couairon2022diffedit} has further refined the landscape of image editing.

Prompt2Prompt \cite{hertz2022prompttoprompt} is designed for image editing within a pre-trained diffusion model by manipulating prompts. It involves injecting a cross-attention map into the diffusion model's sampling process, directing pixels to focus on the tokens of text prompts during diffusion steps to achieve image editing. However, their methodology is confined to altering generated images and does not extend to modifications of real images.

Liu et al. guided a diffusion process using both text and image, creating a composition that closely resembled the provided image and aligned with the given text \cite{kawar2023enhancing}.

Hertz et al. utilized a classifier-free guidance diffusion model to establish a robust anchor point for the reconstruction of the original image through DDIM Inversion. Furthermore, they optimized the Null-text Embedding of the classifier-free guidance diffusion model to achieve precise reconstruction of the original image \cite{mokady2022nulltext}.

A component of our method shares similarities with classifier-free guidance. Our guidance involves making predictions three times: first, utilizing the original image prompt condition embedding; second, using the editing image prompt condition embedding; and third, unconditionally using null-text embedding. Furthermore, for additional precision, we guide the process by calculating differences in the real image domain.

\begin{figure}
    \centering
    \captionsetup{type=figure}
    \includegraphics[width=0.5\textwidth,height=5.5cm]{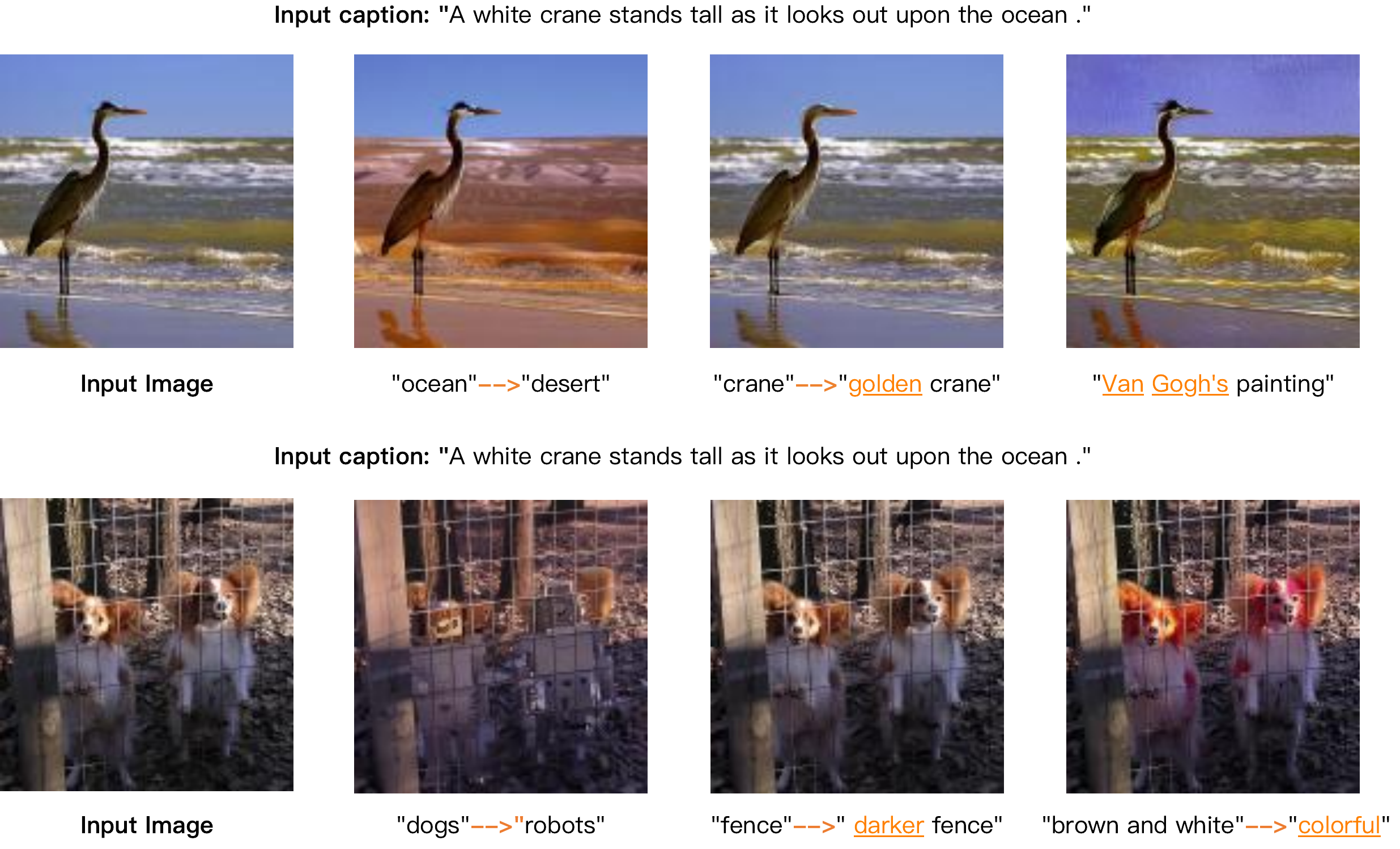}
    
    \captionof{figure}{\textbf{Real image editing using our method.}  It can be seen that our method retains the similarity to the original image while completing the editing.}
\end{figure}%


%% file: sec/3_finalcopy.tex
\section{Method}
\label{sec:formatting}

Let an wanted image that can be described using the edited prompt $P_{edi}$   . Our goal is to edit the real image $X_{src}$ , which can be described using the source prompt $P_{src}$ .We use the edited prompt  $P_{edi}$   , null-text prompt $P_{null}$   , source prompt $P_{src}$ , and perceptual similarity to get the edited image $X_{edi}$  .
We use the Stable Diffusion model \cite{rombach2022highresolution}, where the diffusion forward process is applied to the latent space, using the image encoding $Z_{src} = E (X_{src})$ at the beginning to transfer image to latent space, and the image decoder $X_{edi} = D (Z_{edi})$ at the end of the diffusion backward process to transfer data from latent space to image.
We will apply the guidance on $Z_{src}$ , the hidden space of the real image, and $Z_{edi}$ , the hidden space of the edited image. In the pixel domain, it is guided by calculating the perceived similarity \cite{zhang2018unreasonable} of $X_{src}$ and $X_{edi}$ .

\subsection{Text guidance optimization}
To achieve better text guidance in latent Spaces, we set $E_{null}$  = ψ ("") be the embedding of a null text \cite{ho2022classifierfree}, $E_{src}$ = ψ ("$P_{src}$") be the embedding of source prompt, $Eedi$ = ψ ("$P_{edi}$") be the embedding of edited prompt, and let γ and β be the guidance scale parameters in Equation (1) and (2). The network $\varepsilon_\theta$  is trained to predict artificial noise, and $t$ is the step of diffusion.
We use $noise_{cond}$  to represent the guidance between source prompt predicts noise and null text predicts noise. 
We use $noise_{pred}$  to represent the final guidance noise of this step’s image editing. The guidance prediction is defined by:
\begin{equation}\label{eqn-1} 
\begin{aligned}
&noise_{cond} = \varepsilon_\theta ( Z_t, E_{null},t ) + \gamma *[\varepsilon_\theta ( Z_t ,E_{src},t ) - \\ 
& \varepsilon_\theta ( Z_t , E_{null},t )]
\end{aligned}
\end{equation}		

\begin{equation}\label{eqn-1} 
noise_{pred} = noise_{cond} + \beta *[\varepsilon_\theta ( Z_t , E_{edi},t ) - \varepsilon_\theta ( Z_t, E_{src},t )]					     	 					\end{equation}	

The guidance prediction can also defined by:
\begin{equation}\label{eqn-1} 
\begin{aligned}
&noise_{cond} = \varepsilon_\theta ( Z_t , E_{null}  ,t ) + \gamma *[\varepsilon_\theta ( Z_t , E_{src}  ,t ) - \\ & \varepsilon_\theta ( Z_t ,E_{null}  ,t )]				      \end{aligned}							
\end{equation}	

\begin{equation}\label{eqn-1} 
noise_{pred} = noise_{cond} + \beta *[\varepsilon_\theta ( Z_t , E_{edi}  ,t ) - \varepsilon_\theta ( Z_t , E_{null}  ,t )]					     			 		\end{equation}	

Then we can use the $noise_{pred}$ , step $t$, and latent $Z_t$ at step $t$ to calculate the latent $Z_{t-1}$ at the next step $t -1$.
\begin{figure}
    \centering
    \captionsetup{type=figure}
    \includegraphics[width=0.47\textwidth,height=5.5cm]{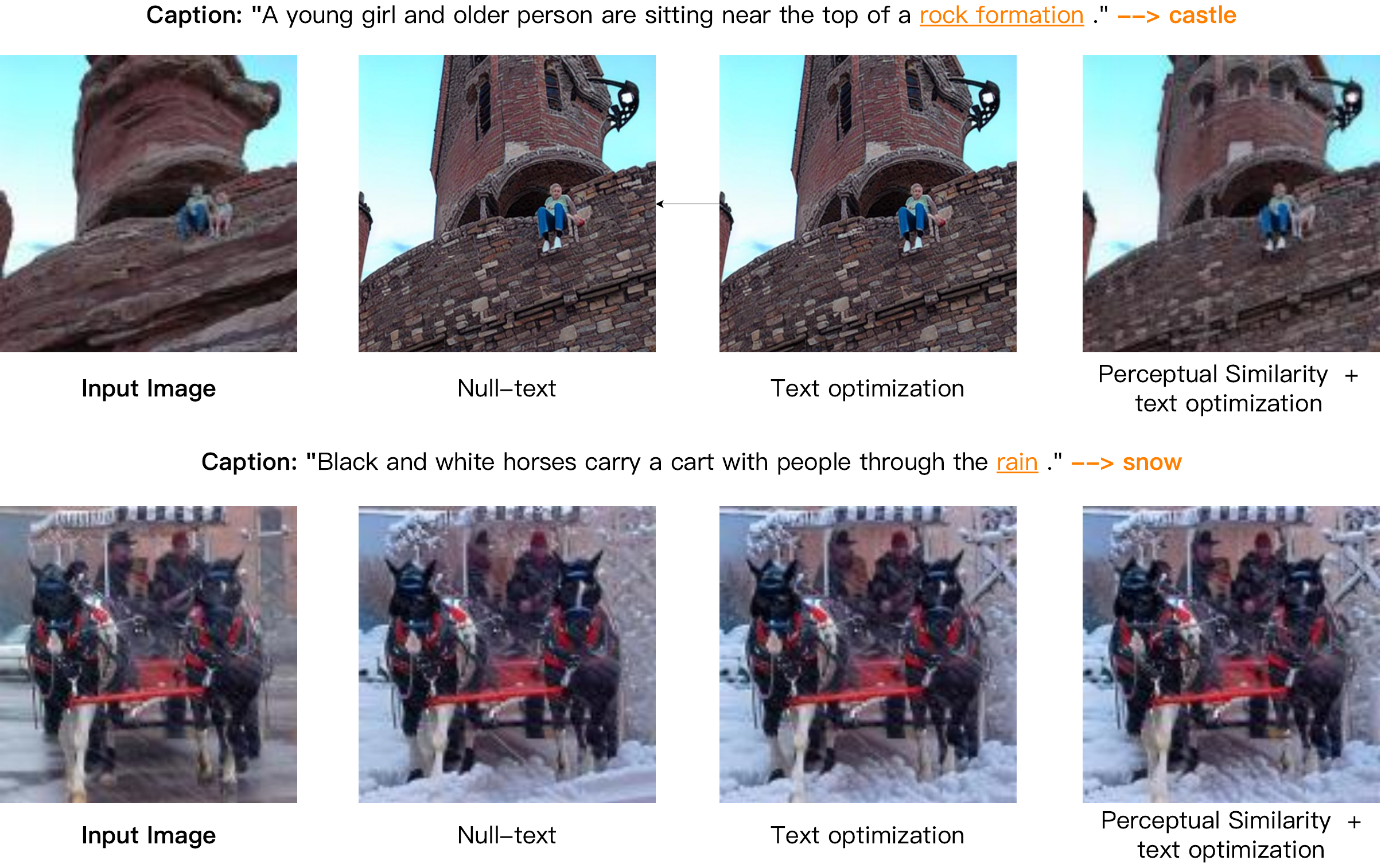}
    
    \captionof{figure}{\textbf{Real image editing under different guidance.}   As evident, despite the minimal distinction between Text optimization image and null-text image(given the utilization of our guidance solely in the initial 20 diffusion steps), the Perceptual Similarity image preserves more details with the original.You can see the Perceptual Similarity + text optimization is more similar to the blur level of the original image, the other images are too clear.}
\end{figure}%

\subsection{Perceptual Similarity guidance}
Now that we have both $Z_t$ and $noise_{pred}$ , we can roughly calculate the $Z_0$ of the edit image latent space at $t$ = 0 :

In stable diffusion \cite{rombach2022highresolution}, $\bar{a_t}$ is the fixed diffusion rate at time $t$ , and we have:

\begin{equation}\label{eqn-1} 
Z_t = \sqrt{\bar{a_t}}Z_0  + \sqrt{1-\bar{a_t}}noise_{pred} ,  noise_{pred} \sim N (0,1)
\end{equation}	

Thus,we can get $Z_0$ by:

\begin{equation}\label{eqn-1} 
Z_0 = \frac{Z_t-\sqrt{1-\bar{a_t}}noise_{pred}}{\sqrt{\bar{a_t}}}   ,  noise_{pred} \sim N (0,1)
\end{equation}	

In order to solve the problem that guidance at the latent space is not accurate enough, we convert the image from the latent space $Z_0$ back to the image domain $X_0$ at every step $t$ .

Then we calculate Perceptual Similarity between the real image $X_{src}$ and our $X_0$ to optimize $Z_t$ in the latent space. Perceptual Similarity is used to evaluate the distance between image patches. Lower means more similar.

%% file: sec/4_Experiment.tex
\section{Experiment}
\label{sec:Experiment}
Our method is designed to focus on intuitive editing, and we gauge fidelity to the original image using LPIPS perceptual distance \cite{zhang2018unreasonable} (lower values are preferable) and PSPR (higher values are preferable). Additionally, we assess fidelity to the target text using CLIP similarity \cite{hessel2022clipscore} (higher values are preferable) on the Flickr8k Dataset.

\begin{table}[!t]
  \centering
  \small
  \begin{tabular}{c c c c}
    \toprule
    Guidance & PSNR &  LPIPS & CLIPScore\\
    \midrule
    Null-text & \textbf{20.2558} & \textbf{0.1857} & 28.5781 \\
    Text optimization & 20.0173 & 0.1919 & \underline{30.4297} \\
    Perceptual Similarity + \\Text optimization & \underline{20.1249} & \underline{0.1894} & \textbf{30.5703}\\
    \bottomrule
  \end{tabular}
  \caption{Comparison of different methods}
  \label{tab:example}
\end{table}

CLIPScore is employed to assess the alignment between the text and the generated image. Notably, the combination of Perceptual Similarity and text optimization exhibits a significant improvement in CLIPScore, indicating a better match between our images and the descriptive text. 

PSNR (Peak Signal-to-Noise Ratio) is utilized to quantify the disparity between the original image and the processed image. The PSNR values across all three experimental groups hover around 20, suggesting that discernible differences are perceptible to the human eye. 

A lower LPIPS (Learned Perceptual Image Patch Similarity)value signifies greater overall similarity between two images. The comparable LPIPS values in the three experimental groups suggest minimal overall differences among the three sets of images.

Furthermore, we conducted evaluations with several individuals, and the results revealed distinctions in details among the three sets of images, with Perceptual Similarity + text optimization emerging as the most effective in preserving the details of the original image.

%% file: sec/5_futurework.tex
\section{Future works}
\label{Future works}
While our method consistently yields satisfactory editing results, there remain areas open for improvement. Our guidance in the image domain relies on a direct comparison between the original and generated images. However, since this comparison involves two complete images, extensive editing may occasionally result in distortions. Although many cases can be rectified by adjusting the image size through guided steps, for future enhancements, we may explore a more focused comparison that highlights unplanned changes in the image.

Moreover, additional limitations stem from the utilization of Stable Diffusion  and Prompt-to-Prompt editing. In certain instances, the text attention map fails to align accurately with the corresponding area in the image. Put differently, the area of the image that we edit text to change will occasionally be chosen incorrectly. This observation underscores a potential pathway for refining our methodology..

\begin{figure}
    \centering
    \captionsetup{type=figure}
    \includegraphics[width=0.47\textwidth,height=5.5cm]{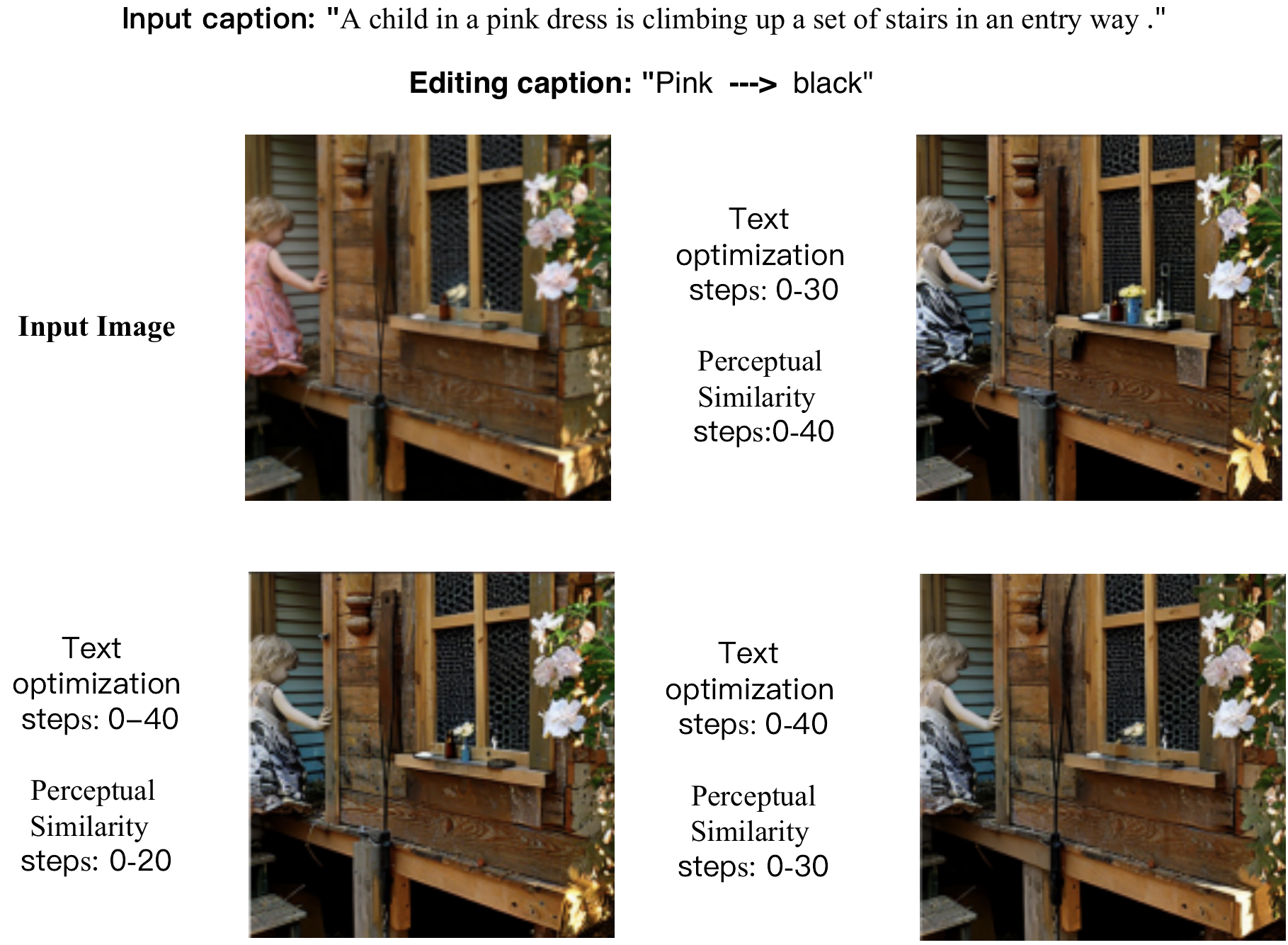}
    
    \captionof{figure}{\textbf{Use our guidance on different diffusion steps.}  Evidently, since the two guidance methods are applied in distinct step ranges, there is a notable variation in the degree of detail restoration from the original image. To illustrate, the restoration of light and shadow in the lower right corner of the example image is excessive in the image domain, resulting in an error where the light and shadow are restored as flowers.}
\end{figure}%